%% file: paper_1.tex
\documentclass[11pt]{article}

\usepackage{acl}

\usepackage{times}
\usepackage{latexsym}
\usepackage[T1]{fontenc}
\usepackage[utf8]{inputenc}
\usepackage{microtype}

\usepackage{graphicx}
\usepackage{amsmath}
\usepackage{amssymb}
\usepackage{booktabs}
\usepackage{array}
\usepackage{xcolor}
\usepackage{url}
\usepackage{tikz}
\usetikzlibrary{arrows.meta, positioning, calc, shadows}
\usepackage{pgfplots}
\pgfplotsset{compat=1.18}
\usepackage{hyperref}
\makeatletter
\g@addto@macro\UrlBreaks{\do\a\do\b\do\c\do\d\do\e\do\f\do\g\do\h\do\i\do\j\do\k\do\l\do\m\do\n\do\o\do\p\do\q\do\r\do\s\do\t\do\u\do\v\do\w\do\x\do\y\do\z\do\A\do\B\do\C\do\D\do\E\do\F\do\G\do\H\do\I\do\J\do\K\do\L\do\M\do\N\do\O\do\P\do\Q\do\R\do\S\do\T\do\U\do\V\do\W\do\X\do\Y\do\Z\do\0\do\1\do\2\do\3\do\4\do\5\do\6\do\7\do\8\do\9}
\makeatother

\title{Agents Don't Paginate: \\
       First-Chunk Selection for LLM Tool Responses}

\author{
  Tatiana Petrova, Andrei Mazniak, Radu State \\
  SnT, University of Luxembourg \\
  {\footnotesize\texttt{\{tatiana.petrova, andrei.mazniak\}@uni.lu}}
}

\begin{document}
\maketitle

\begin{abstract}
Coding agents built on large language models (LLMs), such as Claude
Code, Cursor, OpenAI Codex, GitHub Copilot, and Aider, receive tool
responses that routinely exceed the agent's per-turn token budget. The
standard remedy, pagination, is available in every protocol that
produced these responses; yet across the corpus of session logs from a
public Model Context Protocol middleware we observed no agent-initiated
requests for a second chunk. The first chunk is what the agent reads,
so we ask how often the gold item (the one the agent needs)
is placed \emph{first} in it: the precision-at-1 rate $p_1$.

In a controlled offline benchmark we treat first-chunk selection as a
0/1 knapsack and compare six value functions on 500 SWE-bench Verified
tasks, then test whether $p_1$ matters with a single-turn
file-localisation probe on five language models (4{,}800 LLM calls; not
an end-to-end resolve-rate test). Two pre-registered hypotheses did not
hold and are our main findings. The central one is negative: raising
$p_1$ does not systematically raise downstream accuracy. Per-model
deltas stay under three percentage points (p.p.), are not consistently
signed, and no model is significant; the agent recovers the gold from
anywhere in the chunk, so what reaches its answer is first-chunk
\emph{inclusion}, not the gold's \emph{rank} within it. The second:
adding four file-metadata signals to a keyword scorer hurts $p_1$ by
4.8~p.p.\ (paired significance test, $p = 0.001$).

A parameter-free keyword scorer does raise $p_1$, from a 24.2\%
baseline to 35.0\% ($+10.8$~p.p., far beyond chance;
$p = 3.9\times10^{-8}$), and to 35.8\% with a fallback to the tool's
native ordering when no keyword matches. But by our central finding this is a rank-1 gain, and rank-1
is the part that does not reach the agent's answer: downstream accuracy
does not move.
\end{abstract}

\section{Introduction}
\label{sec:intro}

In a tool-using coding agent, the binding constraint is often not the
model's context length but the ordering of concrete tool-response
items: the file paths returned by a search, the issues returned by a
tracker, or the merge requests returned by a code-review query.
A single turn may surface 50 or more file paths or 200 or more issues,
and in production telemetry from a public Model Context Protocol (MCP)
middleware, 37\% of \texttt{get\_epics} and 28\% of
\texttt{get\_merge\_request\_diffs} calls exceed an 8\,K-token budget.
Every protocol here exposes pagination as the obvious remedy.

Yet the agents do not use it: across the full corpus of session logs
the rate of agent-initiated second-chunk requests is exactly zero, and
after a truncated response the agent moves directly to its next
reasoning or editing step. The structured analyses use a derived
dataset of Bash file-search gold-selection events from these logs
(Section~\ref{sec:setting}); Appendix~\ref{app:cross-corpus} reports
replication on two independent corpora contributed by external teams.

This regularity reframes pre-LLM compression from a pagination problem
to a \emph{selection} problem: the first chunk must contain the item
the agent will use next. We measure first-chunk placement with $p_1$,
the precision-at-1 rate: given $N$ candidate items and a per-response
budget $B$, the fraction of tasks for which the gold item (the issue,
file, or snippet the agent's downstream reasoning eventually cites) is
the top-ranked item in the compressed first chunk. Whether the gold is
merely \emph{included} anywhere in the chunk is a looser, separate
quantity (bounded at $55.8\%$ by the agent's own grep retrieval;
Section~\ref{sec:lessons}). Throughout, $p_1$ is an accuracy rate
(higher is better); a separate symbol $p$ marks statistical
significance (smaller is stronger).

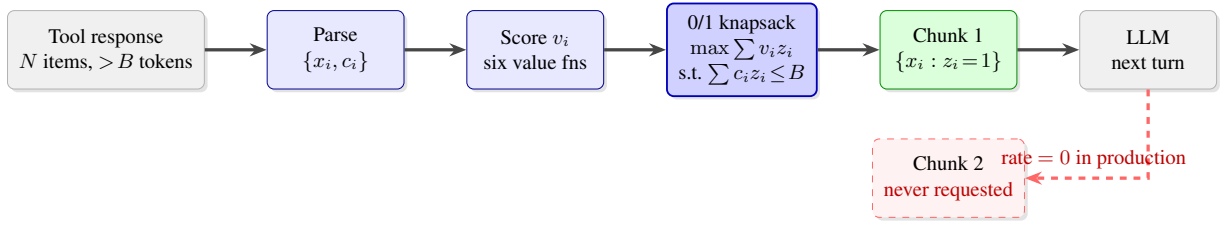
\begin{figure*}[t]
\centering
\input{figures/fig1-pipeline.tex}
\caption{First-chunk selection pipeline. A tool response of $N$
candidate items at cost $c_i$ each is parsed, scored by one of six
value functions $v_i$, and reduced to a first-chunk subset
$\{x_i : z_i = 1\}$ by a 0/1 knapsack under a per-response budget
$B$. The agent never requests chunk~2 in production telemetry
(Section~\ref{sec:setting}), so the selection target is the precision-at-1 rate
$p_1 = \mathbb{P}(x^\star \text{ ranked first})$.}
\label{fig:pipeline}
\end{figure*}

We cast the problem as a 0/1 knapsack (Figure~\ref{fig:pipeline}) and
evaluate six value functions on the 500-task SWE-bench Verified
benchmark in two stages: an algorithmic $p_1$ sweep with no LLM, then a
100-task downstream probe across five language models
(Sections~\ref{sec:method}--\ref{sec:setup}).

We report three findings (Sections~\ref{sec:results}--\ref{sec:lessons}):
(i) the central, negative result: raising the top-1 rate $p_1$ does
not propagate to downstream accuracy (H3), because what binds is
whether the gold reaches the first chunk at all, not its rank within
it; (ii) a parameter-free keyword ranker with a safety fallback does
improve $p_1$ over the production baseline (H1), yet adding four
file-metadata signals to it reverses the gain (H2); and (iii) the
baseline operating point and keyword dominance replicate on two
independent production corpora (Appendix~\ref{app:cross-corpus}).

Code, aggregates, anonymized telemetry, and a Docker image are
released at \url{https://github.com/meteora-pro/devboy-tools}
(Appendix~\ref{app:repro}); the pipeline reproduces on one workstation
with a 24\,GB GPU.

\input{02-setting.tex}
\input{03-method.tex}
\input{04-experimental-setup.tex}
\input{05-results.tex}
\input{07-lessons.tex}
\input{08-related-work.tex}

\input{10-limitations.tex}
\input{11-ethics.tex}

\bibliography{references}

\appendix
\input{appendix.tex}

\end{document}

%% file: figures/fig1-pipeline.tex
\resizebox{\textwidth}{!}{%
\begin{tikzpicture}[
  node distance = 0.7cm and 0.95cm, font=\small,
  box/.style  = {draw, rounded corners=3pt, align=center, inner sep=5pt,
                 minimum height=12mm, minimum width=21mm,
                 drop shadow={opacity=0.22, shadow xshift=0.5mm, shadow yshift=-0.5mm}},
  inp/.style  = {box, fill=gray!12, draw=gray!60},
  proc/.style = {box, fill=blue!9, draw=blue!55!black},
  key/.style  = {box, fill=blue!18, draw=blue!72!black, thick},
  outbox/.style = {box, fill=green!14, draw=green!55!black},
  arr/.style  = {-{Stealth[length=2.6mm,width=2mm]}, line width=0.5mm, gray!55!black},
]
\node[inp] (tool) {Tool response\\[1pt]\footnotesize $N$ items, $>\!B$ tokens};
\node[proc, right=of tool] (parse) {Parse\\[1pt]\footnotesize $\{x_i,c_i\}$};
\node[proc, right=of parse] (score) {Score $v_i$\\[1pt]\footnotesize six value fns};
\node[key, right=of score] (knap) {0/1 knapsack\\[1pt]\footnotesize $\max\sum v_iz_i$\\[1pt]\footnotesize s.t.\ $\sum c_iz_i\!\le\!B$};
\node[outbox, right=of knap] (chunk) {Chunk 1\\[1pt]\footnotesize $\{x_i:z_i\!=\!1\}$};
\node[inp, right=of chunk] (llm) {LLM\\[1pt]\footnotesize next turn};
\foreach \a/\b in {tool/parse,parse/score,score/knap,knap/chunk,chunk/llm}
  \draw[arr] (\a)--(\b);
\node[box, dashed, draw=red!60, fill=red!5, below=0.75cm of chunk, minimum width=21mm] (c2)
  {\footnotesize Chunk 2\\[1pt]\footnotesize \textcolor{red!75!black}{never requested}};
\draw[arr, dashed, red!60] (llm.south) |- (c2.east)
  node[pos=0.72, above, font=\footnotesize, text=red!75!black] {rate $=0$ in production};
\end{tikzpicture}}

%% file: 02-setting.tex
\section{Setting and Baselines}
\label{sec:setting}

We deploy the compression pipeline studied in this paper inside a
publicly released MCP middleware that serves a Claude Code / Cursor /
Codex CLI / Aider client ecosystem. The middleware is distributed as an
\texttt{npm} CLI, an MCP server, and an agent-skill bundle; it federates
seven software-as-a-service (SaaS) providers (a code host, a code-review host, two project
trackers, a meeting transcript service, a knowledge base, and a chat
platform). The pipeline studied here sits between the tool result and
the LLM client and rewrites the tool result body to fit within the
agent's per-turn token budget.

\paragraph{Empirical pagination behaviour.}
Across all over-budget tool responses in our production session logs,
we observed no agent-initiated calls requesting a second chunk with an
explicit cursor; the pagination rate is zero across the corpus. After
a truncated response the agent produces its next turn directly. This
empirical regularity is consistent across coding-agent frontends and
across the seven SaaS providers federated by the middleware. The
structured analysis in subsequent sections is performed on 4{,}175 Bash
file-search gold-selection events extracted from the same logs by an
anonymizing pipeline.

\paragraph{Candidate-list distribution (SWE-bench Verified).}
The benchmark used in Section~\ref{sec:setup} yields a bimodal
distribution: 57.6\% of tasks have $\leq 5$ candidates, 31.4\% sit at
or near the 50-candidate generator cap, and 11.0\% fall in between
(full breakdown in Appendix~\ref{app:bucket-dist}).

\paragraph{Baseline.}
The relevant baseline for file-search responses is the order returned
by the underlying CLI tool (\texttt{grep -rln}, \texttt{rg -l}), which
is the command used by all coding-agent implementations we examined.
This order is filesystem-traversal order (depth-first, lexicographic at
each level), with no relevance ranking. We refer to it as
\textbf{FIFO} (first-in, first-out) throughout. It is also the baseline operating point of
the deployed pipeline: under default tooling, the gold file is the top-ranked candidate in
about one task in four ($p_1 = 24.2\%$). The scorer comparison itself is an
offline SWE-bench study (Section~\ref{sec:setup}); the telemetry
establishes only the pagination-zero regularity and this FIFO
operating point (evidence by source in Appendix~\ref{app:evidence}).

%% file: 03-method.tex
\section{Method}
\label{sec:method}

\subsection{Knapsack formulation}
\label{sec:method:knapsack}

A tool response yields a list of $N$ candidate items $\{x_1, \dots,
x_N\}$, each with a tokenized cost $c_i = |\mathrm{tokenize}(x_i)|$ and a
value $v_i \in \mathbb{R}_{\ge 0}$ assigned by a \emph{value function}
(Section~\ref{sec:method:scorers}). The first-chunk compression problem
is
\begin{equation}
\max_{z \in \{0,1\}^N} \sum_{i=1}^N v_i z_i
\quad \text{s.t.} \quad
\sum_{i=1}^N c_i z_i \le B,
\end{equation}
where $B$ is the budget allocated to this tool response (typically a
small fraction of the agent's total context budget). This is the
canonical 0/1 knapsack. We solve it greedily by density $v_i / c_i$,
which is $1/2$-optimal for the 0/1 knapsack \citep{kellerer2004knapsack}
and runs in $O(N \log N)$. Exact dynamic programming yields no
measurable improvement on the candidate-list sizes we observe
(Section~\ref{sec:setting}).

We define the primary metric
\begin{equation}
p_1 = \mathbb{E}_{\text{task}}\bigl[\,
  \mathbb{1}[\,x^\star = \arg\max\nolimits_i v_i\,]\,\bigr],
\end{equation}
the precision-at-1 rate: the probability that the gold item $x^\star$
(the issue, file, merge request, or snippet the agent's next turn
ultimately cites) is the \emph{top-ranked} item the value function surfaces. We
headline $p_1$ because the agent reads the first chunk from the top.
In our regime $p_1$ and the deeper cutoffs $p_3, p_5, p_{10}$ are all
budget-invariant (the budget admits at least the top ten candidates
even at 1\,K), so we report one value per strategy.

\subsection{Items, cost, value}

For file-search responses (\texttt{grep}, \texttt{rg},
\texttt{Glob}), an item is a single file path. Cost is the token cost of
the path string under the receiving model's tokenizer; values are
scorer-specific. For richer tool responses (e.g., paginated issue
listings), the same formulation applies with composite items (one
issue $\approx$ a multi-field record); we focus on the file-search case
in this paper because it has a clean SWE-bench-derived ground truth
(the gold-modified file).

A \emph{value function} takes the search query (the agent's current
intent, distilled to a token bag), the item, and any per-item metadata
(file extension, modification time, directory depth) and returns a
non-negative score. We evaluate six.

\subsection{Six value functions}
\label{sec:method:scorers}

\paragraph{FIFO} (production baseline). $v_i = N - i$: items keep CLI
order. No query signal; no item-metadata signal.

\paragraph{Random} (variance control). $v_i \sim \mathrm{Uniform}[0, 1]$
i.i.d. Establishes a lower bound on the \emph{no-signal} regime.

\paragraph{Reversed} (adversarial). $v_i = i$: mirror of FIFO. An upper
bound on how badly an actively wrong heuristic damages $p_1$.

\paragraph{Priority-KW} (keyword overlap).
$v_i = \cos(\mathrm{tok}(x_i),\, \mathrm{tok}(q))$, the cosine
similarity between the tokenized item path and the tokenized query in a shared
bag-of-tokens vocabulary built from sub-word splits on
\texttt{[/\_.-]}. No learned weights. No external corpus.

\paragraph{Priority-ALL} (composite).
$v_i = \alpha \cdot \mathrm{KW}(x_i)
      + \beta  \cdot \mathrm{depth\_prior}(x_i)
      + \gamma \cdot \mathrm{ext\_prior}(x_i)
      + \delta \cdot \mathrm{recency}(x_i)
      + \epsilon \cdot \mathrm{filename\_match}(x_i)$.
Depth prior penalises deep paths (test files at depth $\geq 4$ are less
likely to be the gold target); \texttt{ext\_prior} weights canonical
extensions for the language of the repository; \texttt{recency} uses
file modification time; and \texttt{filename\_match} boosts exact
substring matches of any query token against the basename. Coefficients
were chosen \emph{a priori} by hand without per-corpus tuning; we treat
Priority-ALL as the natural first-attempt composite an engineer would
write, not as a learned baseline.

\paragraph{Priority-KW\textsuperscript{+}} (Priority-KW with FIFO safety
fallback).
$v_i = \mathrm{KW}(x_i)$ when $\max_j \mathrm{KW}(x_j) > 0$, else
$v_i = N - i$. The fallback addresses a degenerate corner case: in
approximately 14\% of SWE-bench Verified tasks all candidate keyword
scores are zero: the issue shares no sub-token with any candidate
path (worked example in Appendix~\ref{app:empty-keyword}). Without the
fallback the greedy solver returns the empty set on these tasks and
the LLM receives no candidates.

\paragraph{Why these six value functions.}
The set spans the engineer's design space rather than being
exhaustive, so any change in $p_1$ traces to one design choice. FIFO anchors the production baseline; Random and Reversed bound
the no-signal and adversarial extremes; Priority-KW is the simplest
query-driven signal; Priority-ALL is the natural composite an engineer
would write next; and Priority-KW\textsuperscript{+} adds a single
safety fallback for the empty-score corner case.

%% file: 04-experimental-setup.tex
\section{Experimental Setup}
\label{sec:setup}

We evaluate the six value functions on SWE-bench Verified
\citep{jimenez2024swebench,openai2024verified}, the human-validated
500-task subset of SWE-bench drawn from 12 popular Python repositories.
We chose this benchmark for three reasons: (i) ground-truth gold files
exist per task (the files modified in the human-reference patch), so
$p_1$ is well-defined and not a proxy; (ii) the task corpus spans a
broad range of candidate-list sizes (Section~\ref{sec:setting} reports
the distribution); (iii) the tasks are drawn from real GitHub issues,
so the queries are noisy in the way production queries are noisy. The
Python-only restriction is a limitation we discuss in the Limitations
section.

\paragraph{Candidate generation.}
For each task we run a deterministic grep-based candidate generator that
simulates the file-search behaviour of a coding agent in the absence of
any compression: tokenize the issue text on \texttt{[/\_.-\textbackslash s]},
build a query bag, and emit the union of \texttt{grep -rln} results for
each token that appears in at least one file path. The generator
truncates each candidate list at 50 files; the cap is reached by 136 of
500 tasks (27.2\%) and most of those overlap the ``large'' candidate
bucket in Section~\ref{sec:setting}. We cache candidates per task and
re-use them across all strategies, so each strategy operates on
identical inputs. This generator reproduces the agent's own file-search
(its Bash \texttt{grep}/\texttt{rg} calls), so the recall ceiling it
induces is a property of the deployment, not a weak baseline we chose. We isolate the selection step from
this ceiling in Section~\ref{sec:lessons}.

\paragraph{Budgets.}
We evaluate four budget levels (1{,}000, 2{,}000, 4{,}000, and
8{,}000 tokens) chosen to bracket the operating points of practical
agents (typical per-tool-response slice of a 200\,K-token Sonnet
context).

\paragraph{Experiment E1 (algorithmic).}
We compute $p_1$ for each (task, strategy, budget) cell, no LLM in the
loop. $500 \times 6 \times 4 = 12{,}000$ cells. Runs on CPU; total wall
time approximately 2 minutes.

\paragraph{Experiment E2 (LLM accuracy).}
We sample 100 tasks in which the gold file is present in the candidate
set (localisation is only meaningful when the gold is retrievable;
100\% of this subset is winnable vs 55.8\% of the full 500) and run a
downstream file-localisation probe against five language models:
Claude Opus 4.7, Claude Sonnet 4.5 (both via Anthropic Batch API with
\texttt{cache\_control} enabled), GLM-5.1 with thinking budget 2048 (via
z.ai's Anthropic-compatible endpoint), and gemma4:26b and gpt-oss:20b
(both local via Ollama's native Anthropic-compatible endpoint on a
single RTX 3090 with 24\,GB VRAM). For each (task, strategy, budget,
model) cell we present the LLM with the issue text and the compressed
candidate list and ask it to name the file most likely to contain the
relevant code. Accuracy is the indicator that the named file matches
the gold. Each cell is scored from a single deterministic call
(temperature $= 0$).

Strategy coverage is not uniform across models (cloud models on FIFO
$+$ Priority-KW\textsuperscript{+}; GLM-5.1 on FIFO $+$ Priority-KW;
local models on all three), totalling 4{,}800 LLM calls. Full
coverage matrix in Appendix~\ref{app:repro}.

\paragraph{Cost and reproducibility.}
For cloud models we record the actual billed token counts and compute
\$/correct answer; for local models cost is electricity only and
reported as \$0.00. The full E1 $+$ E2 pipeline reproduces from a
single \texttt{make all} target using a pinned Docker image (total
cloud bill in Section~\ref{sec:results:cost}). All anonymized
aggregates and the full reproducibility kit (hardware, environment
variables, per-stage wall-clock, and Makefile targets) are in
Appendix~\ref{app:repro}.

%% file: 05-results.tex
\section{Results}
\label{sec:results}

We report E1 (algorithmic $p_1$), E2 (downstream LLM accuracy), and
E3 (cost per correct answer).

\subsection{Algorithmic first-position rate}
\label{sec:results:algo}

Table~\ref{tab:p1-strategies} reports $p_1$ over the 500-task
evaluation for each value function, with bootstrap 95\% confidence
intervals computed at the task level.

\begin{table}[t]
\centering
\small
\begin{tabular}{lrc}
\toprule
Strategy & $p_1$ & 95\% CI \\
\midrule
Reversed                     & 2.6\%  & [1.5\%, 4.4\%]    \\
Random                       & 22.6\% & [19.2\%, 26.5\%]  \\
FIFO (production baseline)   & 24.2\% & [20.7\%, 28.1\%]  \\
Priority-ALL                 & 30.2\% & [26.3\%, 34.4\%]  \\
Priority-KW                  & 35.0\% & [30.9\%, 39.3\%]  \\
Priority-KW\textsuperscript{+} & \textbf{35.8\%} & [31.7\%, 40.1\%] \\
\bottomrule
\end{tabular}
\caption{First-position rate $p_1$ (precision@1) on 500 SWE-bench
Verified tasks: how often the gold file is the top-ranked candidate.
Bootstrap 95\% confidence intervals; $p_1$ is constant across the four
budgets within each strategy.}
\label{tab:p1-strategies}
\end{table}

The primary hypothesis H1 (Priority-KW improves $p_1$ over FIFO by at
least 10~p.p.) holds at $+10.8$~p.p.\ (paired McNemar
$p = 3.9\times10^{-8}$; 76 of 500 tasks gain rank-1, 22 lose it). The
fallback adds a further 0.8~p.p. The scorer's gain over FIFO is
concentrated at rank~1 and decays with depth (top-1 $+11.6$, top-10
only $+4.0$~p.p.; $p_1$--$p_{10}$ table in
Appendix~\ref{app:inclusion}): it re-ranks the top of the chunk far
more than it pulls new gold into it, which is exactly why its $p_1$
gain does not reach downstream accuracy (Section~\ref{sec:results:llm}).

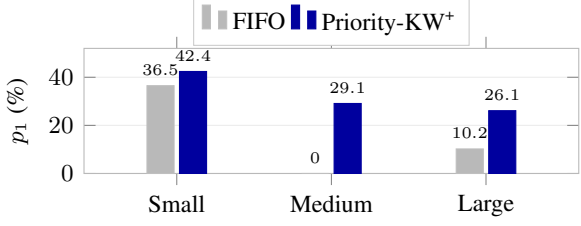
\begin{figure}[t]
\centering
\resizebox{\columnwidth}{!}{%
\begin{tikzpicture}
\begin{axis}[ybar, bar width=10pt, width=8cm, height=3.2cm,
  symbolic x coords={Small,Medium,Large}, xtick=data,
  ymin=0, ymax=52, ylabel={$p_1$ (\%)}, enlarge x limits=0.3,
  ymajorgrids, grid style={gray!15}, axis line style={gray!55},
  ylabel style={font=\small}, tick label style={font=\footnotesize},
  legend style={font=\footnotesize, at={(0.5,1.02)}, anchor=south,
    legend columns=2, draw=gray!40},
  nodes near coords, every node near coord/.append style={font=\tiny,
    /pgf/number format/fixed,/pgf/number format/precision=1}]
\addplot[fill=gray!55,draw=gray!50] coordinates {(Small,36.5)(Medium,0.0)(Large,10.2)};
\addplot[fill=blue!62!black,draw=blue!62!black] coordinates {(Small,42.4)(Medium,29.1)(Large,26.1)};
\legend{FIFO, Priority-KW\textsuperscript{+}}
\end{axis}
\end{tikzpicture}}
\caption{First-position rate $p_1$ (\%) by candidate-list bucket
(Small 1--5, Medium 6--20, Large $\geq$21; task counts $288/55/157$):
FIFO (grey) vs Priority-KW\textsuperscript{+} (blue).}
\label{fig:buckets}
\end{figure}

The average $p_1$ understates the effect on tasks where compression is
the binding constraint. Figure~\ref{fig:buckets} reports $p_1$ by
candidate-list size bucket; the bucket boundaries follow
Section~\ref{sec:setting}.

The lift is largest where FIFO is weakest: the medium and large
buckets, where filesystem-traversal order rarely places the gold first
(Figure~\ref{fig:buckets}).

\subsection{Downstream LLM accuracy}
\label{sec:results:llm}

We ran an offline file-localisation evaluation of five language
models on a 100-task subset of SWE-bench Verified, with per-model
strategy coverage as set out in Section~\ref{sec:setup}. The subset is
drawn from tasks where the gold is reachable in the candidate set
(100\% winnable, vs 55.8\% of the full 500), since localisation is only
meaningful when the gold is present; its per-strategy $p_1$ is
therefore higher than the full set ($0.50/0.70/0.72$ vs
$0.24/0.35/0.36$) and the accuracies below sit near ceiling. Per-model
budget-averaged values are in Appendix~\ref{app:per-bucket-llm}
(Table~\ref{tab:llm-acc}).

\begin{table}[t]
\centering
\small
\begin{tabular}{lrrr}
\toprule
Model & FIFO & keyword & McNemar $p$ \\
\midrule
Claude Opus 4.7   & 94.0 & 93.8 & 1.00 \\
Claude Sonnet 4.5 & 90.8 & 93.0 & 0.25 \\
GLM-5.1           & 93.3 & 90.5 & 0.38 \\
gemma4:26b        & 86.3 & 87.3 & 1.00 \\
gpt-oss:20b       & 81.8 & 82.5 & 1.00 \\
\bottomrule
\end{tabular}
\caption{Downstream accuracy (\%) under FIFO ($p_1 = 0.50$) versus the
keyword scorer ($p_1 = 0.70$--$0.72$; Priority-KW\textsuperscript{+}
where evaluated, else Priority-KW). Accuracies are budget-averaged
(as in Table~\ref{tab:llm-acc}); the McNemar exact test is paired at
the task level (100 tasks, the four near-budget-invariant cells
collapsed by majority vote). Despite a ${\sim}20$~p.p.\ jump in $p_1$,
accuracy moves by at most $\pm 2.8$~p.p.\ and \emph{no} model shows a
significant change.}
\label{tab:h3paired}
\end{table}

Two patterns appear. First, absolute accuracy clusters into three
tiers (frontier $\approx 93\%$, mid, local $\approx 82\%$;
Table~\ref{tab:llm-acc}). Second, \emph{raising $p_1$ does not raise
accuracy.} Table~\ref{tab:h3paired} pairs each model's FIFO cells
against its keyword-scorer cells on the same (task, budget) cells:
moving $p_1$ from $0.50$ to $0.70$--$0.72$ shifts accuracy by between
$-2.8$ and $+2.2$~p.p.; a paired McNemar test at the task level finds
\emph{no} model significant, and the sign is not consistent across
models. The pre-registered H3 (a cell-mean
correlation $\geq 0.85$ between $p_1$ and accuracy) is not supported;
we use these paired tests in place of that correlation, which is
uninformative here (Section~\ref{sec:lessons}).

\paragraph{Conditioning on the gold being ranked first.}
Conditional on $p_1 = 1$ (the gold file is the top-ranked item), the
gemma4--Opus accuracy gap contracts from $7.8$~p.p.\ unconditional to
$1.9$~p.p.\
(per-model conditional accuracies in
Appendix~\ref{app:cond-llm}).

\paragraph{Cross-corpus replication.}
The FIFO operating point ($p_1 = 36.7\%$) and keyword dominance
replicate on two independent production corpora; the benchmark lift and
the pagination-zero claim do not carry over
(Appendix~\ref{app:cross-corpus}).

\subsection{Cost per correct answer}
\label{sec:results:cost}

Table~\ref{tab:cost} reports cost per correct answer derived from the
actual billed token counts for the cloud models and zero for the local
models (electricity only).

\begin{table}[t]
\centering
\small
\begin{tabular}{lrrrr}
\toprule
Model & $n$ calls & Acc. & \$/correct & Cache hit \\
\midrule
gemma4:26b      & 1{,}200 & 86.1\% & \$0.0000 & ---     \\
gpt-oss:20b     & 1{,}200 & 81.1\% & \$0.0000 & ---     \\
GLM-5.1         &   800   & 91.9\% & \$0.0006 & ---     \\
Claude Sonnet 4.5 & 800   & 91.9\% & \$0.0022 & 66.5\%  \\
Claude Opus 4.7   & 800   & 93.9\% & \$0.0307 & 12.6\%  \\
\bottomrule
\end{tabular}
\caption{Aggregate cost over the full E2 run (strategy-averaged
accuracy). Cache hit rates measured under Anthropic Batch API with
default 5-minute time-to-live (TTL).}
\label{tab:cost}
\end{table}

GLM-5.1 matched Sonnet 4.5's strategy-averaged accuracy at roughly
$28\%$ of the per-correct-answer cost; the E2 cloud bill was \$25.09,
of which Opus 4.7 accounted for \$23.05 (key-value (KV) cache and wall-clock detail
in Appendix~\ref{app:repro}).


%% file: 07-lessons.tex
\section{Discussion}
\label{sec:lessons}

Two pre-registered hypotheses (H2, H3) reversed; we report the data
direction. Two operational observations affected how we read the
results.

\subsection{Composite scorers degrade $p_1$ relative to a single signal}

H2 predicted the composite Priority-ALL would beat Priority-KW by
$\geq 3$~p.p.; instead it fell short by 4.8~p.p.\
(Table~\ref{tab:p1-strategies}). Paired on the identical 500
candidate lists, Priority-KW ranks the gold first on 38 tasks where
Priority-ALL does not and loses it on only 14 (McNemar $p = 0.001$),
so the reversal is not a within-noise tie.
The four additional signals in Priority-ALL (depth, extension,
recency, filename) correlate with file \emph{type} and down-weight
correct candidates in atypical workflow locations; the parameter-free
keyword signal is direct evidence about query relevance and dominates
whenever it is present (Appendix~\ref{app:cross-corpus}). The
deployment lesson: measure $p_1$ before adding signals to a working
keyword scorer; more signal is not monotone.

\subsection{Top-1 placement does not predict LLM accuracy}

We pre-registered H3 as a cell-mean correlation
$r(p_1, \text{LLM acc}) \geq 0.85$. That test is uninformative here
because $p_1$ is budget-invariant, so per model it takes only two or
three distinct values and a cell-level $r$ reduces to the sign of a
two-group difference. The paired McNemar tests of
Section~\ref{sec:results:llm} (Table~\ref{tab:h3paired}) test the
question directly and reject H3: the accuracy shift as $p_1$ rises is
not significant for any model and is not consistently signed.

Reading: downstream accuracy depends on whether the gold reaches the
first chunk at all, not on its \textbf{rank within the chunk}. Raising
the top-1 rate $p_1$ leaves accuracy flat because the agent recovers
the gold from anywhere in the chunk it reads; this is reinforced by
the gemma4--Opus accuracy gap collapsing from 7.8 to 1.9~p.p.\ once
the gold is ranked first (Section~\ref{sec:results:llm}). The null is
not a ceiling artefact: the local models on the large bucket sit at
69--76\% accuracy, far from ceiling, yet $p_1$ gains do not help them
(Appendix~\ref{app:per-bucket-llm}). The methodological implication: compression for LLM agents should report
the retrieval metric ($p_1$ or top-$K$) and end-to-end accuracy
separately, and not assume the first tracks the second.

\subsection{Retrieval ceiling versus selection lift}

The grep generator, the agent's own \texttt{grep}/\texttt{rg}
retrieval, misses the gold file in 221 of 500 tasks (44.2\%; e.g.\
\texttt{django-10914}, whose gold \texttt{conf/global\_settings.py}
shares no sub-token with the issue), capping first-chunk
\emph{inclusion} at $55.8\%$. That ceiling is a retrieval limit, not a
selection one. Restricted to the 279 winnable tasks, the keyword scorer
lifts $p_1$ from FIFO's $43.4\%$ to $64.2\%$ ($+20.8$~p.p.;
Appendix~\ref{app:winnable}): selection is a large lever once the gold
is retrievable, and improving retrieval (BM25, dense embeddings) would
raise that ceiling rather than substitute for selection. The same
decomposition explains the flat accuracy. FIFO already places the gold
in the first chunk's top-10 for $86.0\%$ of winnable tasks but at
rank~1 for only $43.4\%$, so the scorer reorders \emph{within} a chunk
that already contains the gold, a reordering the LLM does not need.
Because agents do not paginate, the lever that reaches the answer is
first-chunk inclusion. $p_1$ is a useful retrieval diagnostic, not the
downstream objective.

%% file: 08-related-work.tex
\section{Related Work}
\label{sec:related}

\paragraph{Pre-LLM compression for tool responses.}
Token-level compressors (LLMLingua \citep{jiang2023llmlingua},
LLMLingua-2 \citep{pan2024llmlingua2}, RECOMP \citep{xu2024recomp},
ACON \citep{zhou2024acon}) shorten sequences while preserving task
performance, but operate \emph{after} items are merged into a flat text
stream. First-chunk selection sits one step earlier: a discrete
selection over structural items (paths, records) before rendering, so
the metric is the first-position rate $p_1$, not reconstruction
fidelity.

\paragraph{Retrieval and ranking for code.}
The keyword-overlap scorer is a parameter-free instance of
bag-of-words retrieval. Learned re-rankers (CodeBERT
\citep{feng2020codebert}, UniXcoder \citep{guo2022unixcoder}, dense
retrieval \citep{karpukhin2020dpr}) could replace it but add a learned
component the deployed pipeline does not carry; that is the natural
follow-up.

\paragraph{Tool-use protocol layer.}
First-chunk selection sits in the tool-use protocol layer of MCP
\texttt{ToolAnnotations} \citep{anthropic2025toolannotations},
Anthropic's \emph{Writing Tools for Agents}
\citep{anthropic2026writingtools}, OpenAI function calling
\citep{openai2024functioncalling}, and ResourceLink
\citep{resourcelink2510}; our $p_1$-targeted compression is orthogonal
to all.

\paragraph{Benchmark.}
SWE-bench \citep{jimenez2024swebench} and SWE-bench Verified
\citep{openai2024verified} are the standard benchmarks; we use Verified
as a file-localisation probe, not an end-to-end resolve-rate test.

%% file: 10-limitations.tex
\section*{Limitations}
\label{sec:limitations}

\begin{enumerate}
\item \textbf{Retrieval ceiling.} The deterministic grep candidate
generator caps each list at 50 files and leaves 221 of 500 tasks
(44.2\%) with no candidate matching the gold file under any (strategy,
budget) combination. The absolute $p_1$ ceiling under the present
generator is 55.8\%. A learned retriever (BM25, dense embeddings) is
the natural follow-up; we do not evaluate one here, as it requires
regenerating candidate lists from repository checkouts. We do, however,
separate selection from this ceiling
(Appendix~\ref{app:winnable}): conditional on the gold being
retrievable, the keyword scorer lifts $p_1$ by $+20.8$~p.p.\ over FIFO.

\item \textbf{Benchmark scope.} SWE-bench Verified contains Python-only
tasks drawn from 12 repositories (Django $\approx 46\%$). We do not
evaluate non-Python languages or non-repository corpora.

\item \textbf{Single-turn evaluation.} The downstream LLM accuracy we
measure (Section~\ref{sec:results:llm}) is a single-turn
file-localisation probe, not a full SWE-bench resolve-rate. End-to-end
agent behaviour with patching and test execution is the natural
follow-up.

\item \textbf{Cache TTL artefact.} The Opus 4.7 batch run achieved a
12.6\% KV-cache hit rate under the default 5-minute TTL because batches
exceeded the TTL boundary; a longer TTL is likely to bring it close to
Sonnet 4.5's 66.5\% on the same workload. We have not re-run with
extended TTL.

\item \textbf{Cross-corpus availability.} The side-by-side
replication report (Appendix~\ref{app:cross-corpus}) cannot be
redistributed, and its intent labels were assigned by different,
uncontrolled per-corpus LLM judges, so the external corpora
corroborate the regularity, not a precise effect size.
\end{enumerate}

%% file: 11-ethics.tex
\section*{Ethical Considerations}
\label{sec:ethics}

The production telemetry used in Section~\ref{sec:setting}
and Appendix~\ref{app:cross-corpus} derives from session logs of a public open-source
Model Context Protocol middleware. The anonymising extraction
pipeline emits only aggregate per-event features (shape, token
counts, content-hash prefixes, item counts) and discards raw response
bodies, project identifiers, and user-facing strings at source. The
two external corpora used for cross-corpus replication
(Appendix~\ref{app:cross-corpus}) were contributed by separate teams under
explicit private agreement; we did not see raw logs. SWE-bench
Verified is a public benchmark used unchanged. The reproduction
pipeline costs approximately \$25 in cloud-LLM credits and runs to
completion in roughly 7~hours on a single workstation. We do not see
plausible misuse scenarios specific to this work; the broader risk
surface of LLM-based coding agents is independent of and prior to the
compression layer this paper addresses.

%% file: appendix.tex

\section{LLM File-Localisation Prompt}
\label{app:prompt}

The E2 calls use a stable system prefix (sent with
\texttt{cache\_control: ephemeral} for prompt-cache reuse) plus a
per-task template; both ship with the released code. Abridged system
prefix:

\begin{verbatim}
You are a code-nav assistant.
Given an issue and a truncated
candidate paths (gold inside),
return one JSON line:
  {"chosen_file": <path>,
   "confidence": 0.0-1.0,
   "reasoning": <one sentence>}
Ranking rules, priority order:
 1. keyword match (issue term in
    path) - strongest signal;
 2. source > docs/configs/tests;
 3. shallow paths in source dirs;
 4. named-submodule match;
 5. exact filename > generic;
 6. generic names (utils.py,
    options.py) are risky;
 7. ignore ORDER - not a signal.
\end{verbatim}

\noindent Per-task template (not cached):

\begin{verbatim}
Issue: {{ISSUE_TEXT}}
Candidates ({{N_ITEMS}} total,
to budget={{BUDGET_TOK}}):
{{CANDIDATE_LIST}}
Return the single JSON object.
\end{verbatim}

\noindent Decoding is greedy ($T = 0$); each cell is one call. Rule~7
is deliberate: it probes whether the model honours the paper's claim
that first-chunk \emph{inclusion}, not within-chunk order, drives the
outcome.

\section{Pre-Registration of Hypotheses}
\label{app:preregistration}

Before running E2, we logged five hypotheses with the operating
direction expected at submission time. Two held; three did not. We
list them here verbatim from the project log.

\begin{itemize}
\item \textbf{H1.} Priority-KW improves $p_1$ over the FIFO production
baseline by at least $10$ percentage points averaged across the four
budget levels.
\textsc{Outcome:} pass at $\Delta = +10.8$\,p.p.\
(Section~\ref{sec:results:algo}).

\item \textbf{H2.} The composite scorer Priority-ALL improves $p_1$
over Priority-KW by at least $3$ percentage points.
\textsc{Outcome:} reversed; $\Delta = -4.8$\,p.p.\
(Section~\ref{sec:results:algo}).

\item \textbf{H3.} The cell-mean Pearson correlation between
algorithmic $p_1$ and downstream LLM accuracy across (strategy,
budget) cells is at least $0.85$.
\textsc{Outcome:} not supported. The pre-registered correlation was
uninformative ($p_1$ takes only 2--3 distinct values across cells); a
paired McNemar test per model finds accuracy shifts of $-2.8$ to
$+2.2$~p.p.\ when $p_1$ rises, significant for none of the five
(task-level test; Section~\ref{sec:results:llm}, Table~\ref{tab:h3paired}).

\item \textbf{H4.} Anthropic Batch with \texttt{cache\_control}
reduces input-side cost by at least $40\%$ on identical workloads.
\textsc{Outcome:} not directly tested as stated; we report cache-hit
rates (66.5\% on Sonnet 4.5, 12.6\% on Opus 4.7, default 5-minute
TTL). Cached input tokens are still billed at a reduced rate, so the
hit rate upper-bounds rather than equals the input-cost reduction
(Section~\ref{sec:results:cost}).

\item \textbf{H5.} A 24\,GB-GPU local model reaches accuracy within
$10$\,p.p.\ of Claude Opus 4.7 conditional on the gold being ranked
first ($p_1 = 1$).
\textsc{Outcome:} pass at gap $= 1.9$\,p.p.\ for gemma4:26b
(Section~\ref{sec:results:llm}, Table~\ref{tab:llm-cond}).
\end{itemize}

\section{Per-Bucket LLM Accuracy by Strategy}
\label{app:per-bucket-llm}

Table~\ref{tab:llm-acc} reports file-localisation accuracy per
(model, strategy), averaged over the four budget levels (its
relationship to algorithmic $p_1$ is summarised in the main body as
Table~\ref{tab:h3paired}); Table~\ref{tab:per-bucket-llm} disaggregates
it by candidate-list bucket.

\begin{table}[t]
\centering
\small
\begin{tabular}{llrrr}
\toprule
Tier & Model & FIFO & KW & KW\textsuperscript{+} \\
\midrule
Frontier & Claude Opus 4.7   & 94.0\% & ---     & 93.8\% \\
Mid      & Claude Sonnet 4.5 & 90.8\% & ---     & 93.0\% \\
Mid      & GLM-5.1 (z.ai)    & 93.3\% & 90.5\%  & ---    \\
Local    & gemma4:26b        & 86.3\% & 84.8\%  & 87.3\% \\
Local    & gpt-oss:20b       & 81.8\% & 79.0\%  & 82.5\% \\
\bottomrule
\end{tabular}
\caption{File-localisation accuracy on a 100-task SWE-bench Verified
sub-sample, averaged across four budget levels; per-model strategy
coverage is as in Section~\ref{sec:setup}.}
\label{tab:llm-acc}
\end{table}

\begin{table*}[t]
\centering
\small
\begin{tabular}{llrrrrrr}
\toprule
& & \multicolumn{2}{c}{Small (1--5)} & \multicolumn{2}{c}{Medium (6--20)} & \multicolumn{2}{c}{Large ($\geq 21$)} \\
\cmidrule(lr){3-4}\cmidrule(lr){5-6}\cmidrule(lr){7-8}
Model & Strategy & Acc.\ & $n$ & Acc.\ & $n$ & Acc.\ & $n$ \\
\midrule
Claude Opus 4.7   & FIFO                  & 97.3\% & 56 & 100.0\% & 8 & 87.5\% & 36 \\
Claude Opus 4.7   & Priority-KW\textsuperscript{+} & 96.4\% & 56 & 100.0\% & 8 & 88.2\% & 36 \\
Claude Sonnet 4.5 & FIFO                  & 98.2\% & 56 & 87.5\%  & 8 & 79.9\% & 36 \\
Claude Sonnet 4.5 & Priority-KW\textsuperscript{+} & 98.2\% & 56 & 100.0\% & 8 & 83.3\% & 36 \\
GLM-5.1           & FIFO                  & 99.1\% & 56 & 87.5\%  & 8 & 85.4\% & 36 \\
GLM-5.1           & Priority-KW           & 95.1\% & 56 & 90.6\%  & 8 & 83.3\% & 36 \\
gemma4:26b        & FIFO                  & 93.3\% & 56 & 84.4\%  & 8 & 75.7\% & 36 \\
gemma4:26b        & Priority-KW           & 89.7\% & 56 & 87.5\%  & 8 & 76.4\% & 36 \\
gemma4:26b        & Priority-KW\textsuperscript{+} & 96.0\% & 56 & 87.5\%  & 8 & 73.6\% & 36 \\
gpt-oss:20b       & FIFO                  & 91.1\% & 56 & 65.6\%  & 8 & 70.8\% & 36 \\
gpt-oss:20b       & Priority-KW           & 88.4\% & 56 & 59.4\%  & 8 & 68.8\% & 36 \\
gpt-oss:20b       & Priority-KW\textsuperscript{+} & 92.4\% & 56 & 75.0\%  & 8 & 68.8\% & 36 \\
\bottomrule
\end{tabular}
\caption{LLM accuracy by candidate-list bucket and strategy
(budget-averaged). Source: \texttt{swe\_bench\_llm\_by\_bucket.csv}.}
\label{tab:per-bucket-llm}
\end{table*}

\section{Inclusion at Depth ($p_1$ through $p_{10}$)}
\label{app:inclusion}

Table~\ref{tab:inclusion} gives the gold's first-chunk depth at four
cutoffs ($p_K$ = gold among the top-$K$ shown items); like $p_1$, these
are budget-invariant here. The keyword signal raises inclusion at every cutoff,
but its advantage over FIFO is concentrated at rank~1 and decays with
depth ($p_1$ $+11.6$ $\to$ $p_{10}$ $+4.0$~p.p.). This is consistent
with the downstream result (Section~\ref{sec:results:llm}): the scorer
mainly re-orders the top of the chunk rather than changing whether the
gold reaches a multi-item chunk, so it moves $p_1$ far more than the
top-$K$ inclusion the LLM actually consumes.

\begin{table}[t]
\centering
\small
\begin{tabular}{lrrrr}
\toprule
Strategy & $p_1$ & $p_3$ & $p_5$ & $p_{10}$ \\
\midrule
Reversed                       & 2.6  & 10.0 & 12.6 & 14.8 \\
Random                         & 22.6 & 30.2 & 33.8 & 38.4 \\
FIFO                           & 24.2 & 38.8 & 44.0 & 48.0 \\
Priority-ALL                   & 30.2 & 42.0 & 46.6 & 49.4 \\
Priority-KW                    & 35.0 & 44.2 & 47.0 & 49.0 \\
Priority-KW\textsuperscript{+} & 35.8 & 46.4 & 49.8 & 52.0 \\
\bottomrule
\end{tabular}
\caption{First-chunk inclusion at depth (\%): gold among the top-$K$
shown items, budget-invariant
(\texttt{swe\_bench\_strategies.csv}). The keyword scorer's gain over
FIFO shrinks with depth ($p_1$ $+11.6$, $p_3$ $+7.6$, $p_5$ $+5.8$,
$p_{10}$ $+4.0$).}
\label{tab:inclusion}
\end{table}

\section{Selection Lift Conditional on Retrieval}
\label{app:winnable}

To separate the selection step from the candidate generator's recall
ceiling (Section~\ref{sec:lessons}), Table~\ref{tab:winnable} restricts
the algorithmic sweep to the 279 \emph{winnable} tasks (those whose
gold file the grep generator places in the candidate set; the other 221
are unreachable by any selection method). On this subset
Priority-KW\textsuperscript{+} lifts the first-position rate $p_1$ by
$+20.8$~p.p.\ over FIFO ($43.4 \to 64.2$), against the $+11.6$~p.p.\
read off the full 500-task average: the full-set $p_1$ and its
$55.8\%$ ceiling are dominated by the generator's recall, not by the
selection step. Depth-10 inclusion ($p_{10}$: gold anywhere in the
first chunk's top ten) is already high under FIFO ($86.0\%$) and rises
only to $93.2\%$, while $p_1$ rises from $43.4\%$ to $64.2\%$. On
winnable tasks the scorer thus mainly reranks the top of a chunk that
already contains the gold. This is the algorithmic counterpart of the
flat downstream accuracy in Section~\ref{sec:results:llm}: a
$+20.8$~p.p.\ rank-1 gain leaves end-to-end accuracy unmoved because
the gold was already inside the chunk the LLM reads. The 100-task LLM
sub-sample of Section~\ref{sec:setup} is drawn from these 279
(and is slightly easier: per-strategy $p_1$ $0.50/0.70/0.72$).

\begin{table}[t]
\centering
\small
\begin{tabular}{lrr}
\toprule
Strategy & $p_1$ & $p_{10}$ \\
\midrule
FIFO (production baseline)     & 43.4\% & 86.0\% \\
Priority-ALL                   & 54.1\% & 88.5\% \\
Priority-KW                    & 62.7\% & 87.8\% \\
Priority-KW\textsuperscript{+} & \textbf{64.2\%} & \textbf{93.2\%} \\
\bottomrule
\end{tabular}
\caption{First-position rate $p_1$ and depth-10 inclusion $p_{10}$ on
the 279 winnable tasks (gold present in the grep candidate set); both
are budget-invariant. Conditional on retrieval, the keyword scorer
lifts $p_1$ by $+20.8$~p.p.\ over FIFO, but inclusion ($p_{10}$) is
already high under FIFO. Most of the lift is reranking within an
already-including chunk, which is why downstream accuracy is flat.
Source: \texttt{swe\_bench\_strategy\_results.parquet}.}
\label{tab:winnable}
\end{table}

\section{Empty-Keyword Corner Case}
\label{app:empty-keyword}

Approximately 14\% of SWE-bench Verified tasks produce all-zero
keyword scores: every candidate file path shares no sub-token with the
issue text, typically because the issue describes a behaviour while
the gold file is named after an internal class
(e.g.\ \texttt{options.py}, \texttt{base.py}). Without a fallback, the
greedy knapsack admits items in cost order with arbitrary tie-breaks
and the LLM receives no candidates. The Priority-KW\textsuperscript{+}
fallback replaces the all-zero score vector with FIFO order on these
tasks. An illustrative example from the production run:

\begin{quote}\footnotesize
\textbf{Task} \texttt{django-12713} (budget $= 2$\,K, candidates $= 2$).\\
\textbf{Gold file:} \texttt{django/contrib/admin/options.py}.\\
\textbf{FIFO selection:} 2 items, gold at rank 0. $p_1 = 1$ (pass) \\
\textbf{Priority-KW selection:} 0 items (empty set). $p_1 = 0$ (fail) \\
\textbf{Priority-KW\textsuperscript{+} selection:} falls back to FIFO; 2
items, gold at rank 0. $p_1 = 1$ (pass)
\end{quote}

\section{Reproducibility Kit}
\label{app:repro}

\paragraph{Hardware.}
AMD Ryzen 5950X (16 cores), NVIDIA RTX 3090 (24\,GB VRAM), 128\,GB
RAM, 100\,GB free disk. Operating system: Windows 10 with Git Bash and
Docker Desktop.

\paragraph{Software.}
Python 3.12.10; \texttt{uv} 0.11.7 (PEP-723 script runner);
Rust 1.90; Ollama 0.21.0 (native Anthropic-compatible endpoint at
\texttt{/v1/messages}).

\paragraph{Credentials and subscriptions.}
\texttt{ANTHROPIC\_API\_KEY} (prepaid credits, not OAuth tokens);
\texttt{ZAI\_API\_KEY} (GLM Coding Plan, \$3/month subscription;
billed \$0.44 over our run);
\texttt{OLLAMA\_NATIVE\_URL} for local inference;
optional \texttt{HF\_TOKEN} for faster SWE-bench download.

\paragraph{Cost breakdown.}
E1 (algorithmic): 12{,}000 cells in $\sim 2$\,min CPU, \$0.\\
E2 cloud (per model, all on 100 tasks $\times$ 4 budgets $\times$
strategies): Sonnet 4.5 batch $\sim 2$\,min, \$1.59; GLM-5.1
$\sim 210$\,min, \$0.44; Opus 4.7 batch $\sim 2$\,min, \$23.05.\\
E2 local (Ollama): gemma4:26b $\sim 137$\,min, gpt-oss:20b
$\sim 115$\,min, electricity only.\\
Total cloud billed cost: \$25.09; total wall time including local
inference and aggregation: $\sim 7$\,hours.

\paragraph{Reproduction targets.}
Single \texttt{make all} target reproduces from a clean repository:
\texttt{make data} (Stage~1: SWE-bench download + candidate
generation, ${\sim}60$\,min);
\texttt{make e1} (Stage~2: algorithmic sweep,
${\sim}2$\,min);
\texttt{make e2-cloud} (Stages 4--5: cloud LLM evaluation,
${\sim}3.5$\,hrs);
\texttt{make e2-local} (Stage~3: local LLM evaluation,
${\sim}4$\,hrs, GPU required);
\texttt{make aggregate} and \texttt{make plots} (Stage~6, $\sim 15$\,s).

\paragraph{Aggregates released.}
All anonymized aggregates (\texttt{swe\_bench\_strategies.csv},
\texttt{swe\_bench\_strategies\_by\_bucket.csv}, \texttt{swe\_bench\_llm.csv},
\texttt{swe\_bench\_llm\_by\_bucket.csv}, \texttt{swe\_bench\_cost.csv},
\texttt{swe\_bench\_corr.csv}, and the per-task sweep
\texttt{swe\_bench\_strategy\_results.parquet} behind
Appendix~\ref{app:winnable}) and the analysis notebook
\texttt{paper1\_analysis.ipynb} are at \url{https://github.com/meteora-pro/devboy-tools}.
The reproducibility Docker image hash is recorded in the repository.

\section{Candidate-List Distribution and Bucket Sizes}
\label{app:bucket-dist}

Across the 500 SWE-bench Verified tasks, the candidate-list distribution
is bimodal: median 4, 25th percentile 1, with a long tail extending to
the hard cap of 50 candidates per task (set by the deterministic grep
generator in Section~\ref{sec:setup}). Of 500 tasks, 136 (27.2\%) sit
at the cap. The bucketing used in Section~\ref{sec:results} (small,
1--5; medium, 6--20; large, $\geq 21$) partitions the benchmark
into 288, 55, and 157 tasks respectively (57.6\%, 11.0\%, 31.4\%). Of
the large bucket, 86.6\% (136 of 157) sit at the candidate cap.
FIFO's $0\%$ first-position rate on the medium bucket
(Figure~\ref{fig:buckets}) is not an anomaly: for none of those 55
tasks does filesystem-traversal order place the gold first among the
6--20 candidates.

\section{Conditional LLM Accuracy}
\label{app:cond-llm}

Section~\ref{sec:results:llm} reports unconditional accuracy
(Table~\ref{tab:llm-acc}, Appendix~\ref{app:per-bucket-llm}) and a one-sentence
summary of accuracy conditional on the gold file being ranked first.
Table~\ref{tab:llm-cond} below gives the conditional accuracy
per model, on cells where $p_1 = 1$.

\begin{table}[t]
\centering
\small
\begin{tabular}{lrr}
\toprule
Model & Cond.\ acc. & $n$ cells \\
\midrule
Claude Opus 4.7   & 96.7\% & 488 \\
Claude Sonnet 4.5 & 97.5\% & 488 \\
GLM-5.1           & 96.9\% & 480 \\
gemma4:26b        & 94.8\% & 768 \\
gpt-oss:20b       & 93.1\% & 768 \\
\bottomrule
\end{tabular}
\caption{Accuracy conditional on $p_1 = 1$ (gold ranked first). The
unconditional gemma4--Opus gap of $7.8$~p.p.\ contracts to
$1.9$~p.p.\ under this conditioning.}
\label{tab:llm-cond}
\end{table}

\paragraph{Robustness to budget.}
To confirm the H3 null (Section~\ref{sec:results:llm}) is not an
artefact of the near-ceiling winnable subset, we repeat the paired
FIFO-vs-Priority-KW\textsuperscript{+} accuracy comparison per budget,
pooled over the four models with KW\textsuperscript{+} coverage
($n = 400$ pairs each): the paired keyword gain is $+0.8$, $+1.5$,
$+1.0$, and $+0.5$~p.p.\ at the 1\,K, 2\,K, 4\,K, and 8\,K budgets.
Even on the hardest cells (tasks where FIFO does not rank the gold
first, at the tightest 1\,K budget) the gain is only $+4.0$~p.p.\
($79.0\% \to 83.0\%$, $n = 200$). Raising $p_1$ thus does not raise
downstream accuracy even where the budget binds hardest; consistently,
the keyword scorer \emph{without} the FIFO fallback (the \texttt{KW}
column of Table~\ref{tab:llm-acc}) is slightly downstream-negative on
the three models where it was evaluated.

\section{Cross-Corpus Replication (Full Table)}
\label{app:cross-corpus}

To distinguish results that reflect the deployment from those that
reflect the underlying coding-agent workflow, we obtained two
independent Bash file-search gold-selection corpora extracted by
separate teams running the same anonymising pipeline on their own
Claude Code session logs (only aggregate per-event statistics shared,
no raw response bodies).

\begin{table*}[t]
\centering
\small
\begin{tabular}{lrrrr}
\toprule
Corpus & Sessions & Events & FIFO $p_1$ & KW share \\
\midrule
Ours (this paper)      & 2{,}607 & 4{,}175 & 36.7\%      & 83.5\% \\
Corpus~B (independent) &   689   &   590   & 35.4\%      & 80.8\% \\
Corpus~A (independent) &   221   &   145   & 24.1\%$^\dagger$ & 85.4\%$^\ddagger$ \\
\bottomrule
\end{tabular}
\caption{Cross-corpus replication. $^\dagger$Corpus~A reports on the
small (3--9 candidate) bucket; matches our small-3--9 baseline of
24\%. $^\ddagger$Corpus~A's keyword-signal share is on the
real-signal subset only.}
\label{tab:cross-corpus}
\end{table*}

The FIFO baseline reproduces on Corpus~B at $35.4\%$ vs ours at
$36.7\%$ (a 1.3~p.p.\ gap); the keyword-signal share sits in an
80\%--85\% band across all three corpora. The replication scope is
limited to the deployment regularity (FIFO baseline and keyword
dominance); the pagination-zero finding of Section~\ref{sec:setting}
is observed only on our middleware and is not verified
externally. Caveats: Corpus~A and
Corpus~B use different LLM judges for the workflow-intent labelling
step; we did not control for inter-judge variation. Corpus~B's
candidate-list distribution has a longer tail than ours
(99th-percentile $= 89.9$ items, maximum $= 278$). The replication
does not extend the change-of-scorer findings to a different
deployment population but it shows that the baseline and the
dominant keyword signal are not artefacts of one team's workflow.


\section{Evidence Sources}
\label{app:evidence}

Table~\ref{tab:evidence} separates which claims rest on production
telemetry, on the offline SWE-bench Verified study, and on the two
external corpora, so production observations are not read as benchmark
results or vice versa.

\begin{table}[t]
\centering
\small
\begin{tabular}{@{}p{0.24\linewidth}p{0.35\linewidth}p{0.31\linewidth}@{}}
\toprule
Source & Establishes & Scope / caveat \\
\midrule
Production telemetry (this middleware) & pagination-zero; the FIFO operating point ($p_1 = 36.7\%$ on 4{,}175 events) & one deployment; not externally verified \\
SWE-bench Verified (offline) & the scorer comparison ($p_1$), downstream localisation accuracy, the $55.8\%$ grep ceiling & Python-only; single-turn probe, not resolve-rate \\
External corpora B, A & FIFO baseline and keyword dominance replicate (80--85\% keyword share) & aggregate-only; uncontrolled per-corpus judges \\
\bottomrule
\end{tabular}
\caption{Evidence by source. Pagination-zero and the operating point
come from production telemetry; the scorer comparison and downstream
accuracy are an offline SWE-bench study; the external corpora
corroborate the regularity, not the benchmark effect sizes.}
\label{tab:evidence}
\end{table}